\pdfoutput=1
\documentclass[11pt]{article}
\usepackage{siunitx}
\usepackage{amsmath}
\usepackage{graphicx}
\usepackage{dsfont}
\usepackage{booktabs}
\usepackage{array}
\usepackage{booktabs} 
\usepackage{xspace}
\usepackage{arydshln}
\usepackage{longtable} % custom for spanning table over multiple pages
\usepackage[review]{EACL2023}
\usepackage{times}
\usepackage{latexsym}

\usepackage[T1]{fontenc}
\usepackage[utf8]{inputenc}
\usepackage{microtype}
\usepackage{inconsolata}

\title{Encoder-Decoder Framework for Interactive Free Verses with Generation with Controllable High-Quality Rhyming}

\author{
    \begin{tabular}[t]{c}
        \bf Tommaso Pasini\thanks{Huawei London Research Center, UK}  \footnotemark[2] \\ \small \texttt{p.tommaso@gmail.com} \\
        \bf Jinhua Du\footnotemark[1] \\ \small \texttt{jinhua.du@huawei.com} \\
    \end{tabular}
    \begin{tabular}[t]{c}
        \bf Alejo López-Ávila\footnotemark[1]  \thanks{Both authors contributed equally} \\ \small \texttt{alejo.lopez.avila@huawei.com} \\
        \bf Yubing Wang\footnotemark[1] \\ \small \texttt{yubingwang@huawei.com} \\
    \end{tabular}
    \begin{tabular}[t]{c}
        \bf Husam Quteineh\footnotemark[1] \\ \small \texttt{husam.quteineh@huawei.com} \\
        \bf Ze Li\thanks{Noah's Ark Lab-Hong Kong, ZH} \\ \small \texttt{lize23@huawei.com} \\
    \end{tabular}
    \begin{tabular}[t]{c}
        \bf Gerasimos Lampouras\thanks{Noah's Ark Lab-London, UK} \\ \small \texttt{gerasimos.lampouras@huawei.com} \\
        \bf Yusen Sun\thanks{Interactive Media CSDD} \\ \small \texttt{sun.yusen1@huawei.com} \\
    \end{tabular}
}

\newcommand{\wasabi}{Wasabi\xspace}

\newcommand{\tflargertl}{T5-L$_\text{\textsc{rtl}}^\text{Rand}$}
\newcommand{\tflargertlpre}{T5-L$_\text{\textsc{rtl}}^\text{Pretrain}$}
\newcommand{\tflargeplane}{T5-L$^\text{Pretrain}$}
\newcommand{\tfbase}{T5-B$_\text{\textsc{lwf}}$}
\newcommand{\tflarge}{T5-L$_\text{\textsc{lwf}}$}

\newcommand{\tflargepretrain}{T5-L$_\text{\textsc{lwf}}^\text{Pretrain}$}
\newcommand{\tfbaserandom}{T5-B$_\text{\textsc{lwf}}^\text{Rand}$}
\newcommand{\tflargerandom}{T5-L$_\text{\textsc{lwf}}^\text{Rand}$}
\newcommand{\tflwfplus}{T5-L$_\text{\textsc{lwf+epr}}^\text{Pretrain}$}

\begin{document}
\maketitle
\begin{abstract}
Composing poetry or lyrics involves several creative factors, but a challenging aspect of generation is the adherence to a more or less strict metric and rhyming pattern. To address this challenge specifically, previous work on the task has mainly focused on reverse language modeling, which brings the critical selection of each rhyming word to the forefront of each verse. On the other hand, reversing the word order requires that models be trained from scratch with this task-specific goal and cannot take advantage of transfer learning from a Pretrained Language Model (PLM). We propose a novel fine-tuning approach that prepends the rhyming word at the start of each lyric, which allows the critical rhyming decision to be made before the model commits to the content of the lyric (as during reverse language modeling), but maintains compatibility with the word order of regular PLMs as the lyric itself is still generated in left-to-right order. We conducted extensive experiments to compare this fine-tuning against the current state-of-the-art strategies for rhyming, finding that our approach generates more readable text and better rhyming capabilities. Furthermore, we furnish a high-quality dataset in English and 12 other languages, analyse the approach's feasibility in a multilingual context, provide extensive experimental results shedding light on good and bad practices for lyrics generation, and propose metrics to compare methods in the future.
\end{abstract}
% TODO: they require all the libraries in the section of artifacts. We may add a ref to huggingface somewhere

\section{Introduction}
Lyrics generation, the task of generating lyrics based on desiderata defined by a user, e.g., genre or topic, is gaining momentum thanks to the recent advances in text generation. Generating lyrics, however, has its peculiarities, making it a different task from open text generation. 
Indeed, songs need to follow a high-level structure, defining choruses and verse and adhering to rhyming constraints. This is similar to the task of poetry generation, but songs present a vocabulary and styles that differentiate them. We chose Lyrics Generation since lyrics form our datasets and inherit these peculiarities. In general, a few approaches have been proposed either for specific cases, e.g., rap lyrics generation \cite{xue-etal-2021-deeprapper}, or in a more general fashion to adhere with desiderata from English songwriters \cite{ram-etal-2021-saywhat}, or to incorporate verse structure within a model \cite{li-etal-2020-rigid}.

\begin{figure}[t]
\centering
\includegraphics[scale=0.45]{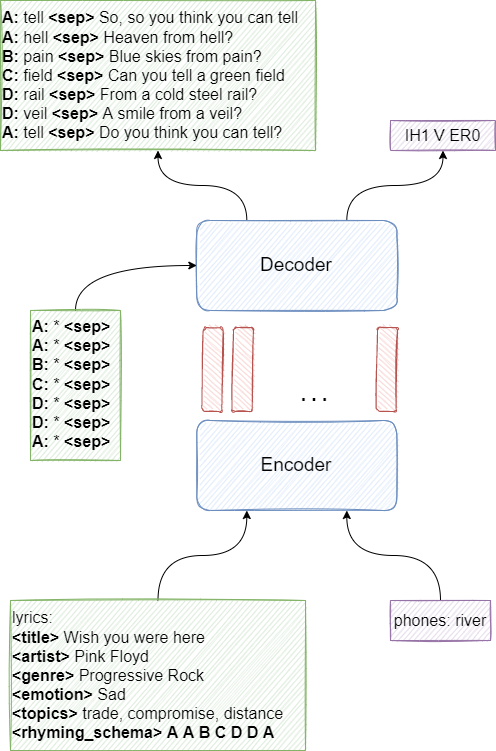}
\caption{Drawing of the proposed model. Green boxes correspond to the main fine-tuning strategy \textsc{LWF}, while the violet ones correspond to the \textsc{LWF+EPR} approach.}
\label{fig:lwf_model}
\end{figure}

This work mainly focuses on the rhyming control aspect of generating lyrics. Previous work on the task \cite{xue-etal-2021-deeprapper, li-etal-2020-rigid} has focused on reverse language modeling, i.e. training a model to generate the output in a right-to-left manner (RTL). This has the benefit of bringing the critical selection of the rhyming word to the forefront of each verse, ensuring that it is unaffected by the semantic context of the verse. The downside of this approach is that reversing the word order requires that models be trained from scratch with this task-specific goal. As such, these approaches cannot take advantage of transfer learning from Pretrained Language Models (PLM) which are generally trained through left-to-right conditional language modeling.

We propose a novel fine-tuning strategy, namely Last Word First (LWF), that can be used for generating human-like text with rhyming control. This strategy fine-tunes a model under a structure where the rhyming word (i.e. the last word of each verse) is prepended at the start of the verse as well (Fig.~\ref{fig:lwf_model}) and constrained (during inference) to follow the rhyming schema. This enables the model to follow a user-defined rhyming pattern and benefit from bringing the critical selection of the rhyming word to the forefront (as in reverse language modeling), while the rest of the verse is still being generated in a left-to-right manner. Our strategy allows us to fine-tune PLMs and benefit from transfer learning, requiring much less data and computing power, where previous work often required retraining a model from scratch. To the best our knowledge, no work has investigated finetuning PLMs to generate lyrics following an arbitrary rhyming pattern defined by a user. 

We additionally condition the lyrics generation on various user-defined aspects, like genre or a specific artist’s style, and explore how this strategy can be augmented through a secondary training objective to predict the Ending Phonetic Representation (EPR) of each word.
Finally, while previous work primarily focused on only one language (e.g. English or Chinese), %which limits the ability to create lyrics in other languages. 
we introduce a novel high-quality dataset of lyrics in English and 12 other languages and use it to demonstrate that LWF is language-agnostic.
%In this paper, we propose a novel fine-tuning approach, shown in Fig.~\ref{fig:lwf_model}, that can be applied to any \PLM to create verses interactively following not only several attributes such as genre, artist's style, song title, and topics but also a given rhyme schema without requiring training the model from scratch, as usual for those approaches relying on reverse language modeling or proposing new models \cite{xue-etal-2021-deeprapper, li-etal-2020-rigid}. Since we do not consider musicality, this method can also be applied to poetry. However, we use datasets of song lyrics, and they have their vocabulary and structure so we will refer to this as lyrics generation. We also provide lyrics data in 13 languages enriched with rhyme schema at the paragraph level to encourage further research in this direction.
%To summarise, 
Our contribution is threefold:
\begin{enumerate}
    \item A novel approach to adapt pretrained models so that they appropriately follow a given rhyming schema, enabling meaningful outputs and precision in rhyming while benefiting from PLM transfer-learning;
    \item We show that our approach outperforms previous SOTA techniques like RTL;
    \item High-quality data in 13 languages for lyrics generation augmented with rhyme schema at the paragraph level;
    \item Extensive experiments and error analysis, showing the pitfalls of current models, including multiple metrics over different aspects of the generation, like diversity (distinct-2, 3 and 4), Mauve, and a new metric on copyright.
\end{enumerate}
% Code and data are available at \link.
Code\footnote{\url{https://github.com/researcher1741/lyrics_generation}} and data\footnote{\url{https://www.kaggle.com/datasets/alejop/lyrics-english-section-dataset-rhyme}}\footnote{\url{https://www.kaggle.com/datasets/alejop/lyrics-multilingual-section-dataset-rhyme}}.

%%%%%%%%%%%%%%%%%%%%%%%%%%%%%%%%%%%%%%%%%%%%%%%%%%%%%%%%%%%%%%%%%%%%%%
%%%%%%%%%%%%%%%%%%%%%%%%%%%%%%%%%%%%%%%%%%%%%%%%%%%%%%%%%%%%%%%%%%%%%%

\section{Related Work}
The task of lyrics generation has to be viewed in the broader context of text generation. Text generation has recently gained much attention thanks to large pretrained language models, e.g., GPT-2 \cite{radford-etal-2019-gpt2}, and GPT-3 \cite{brown-etal-2020-gpt3}, or encoder-decoder models, e.g., BART \cite{lewis-etal-2020-bart}, and T5 \cite{raffel-etal-2020-t5}. Compared to those only trained on the target task, the main advantage of such models is the so-called \textit{knowledge transfer}, i.e., during their pretraining, assimilated knowledge can be later reused in different downstream tasks. A recent challenge in text generation is to add constraints to a model, from soft conditions, such as respecting a given style, to more complex rules, e.g., respecting a predefined schema for the output text, as when writing a poem.

Lyrics generation is a long-standing task in NLP, with the first attempts dating back to the 1960s \cite{100}. More complex systems started spawning around the 2000s \cite{gervas-00-wasp,manurung-04-evolutionary} and recently reached a more satisfactory performance thanks to the advent of deep learning, recurrent neural networks and PLMs \cite{wockener-etal-2021-end,shao-etal-21-sentiment,li-etal-2020-rigid,ormazabal-etal-2022-poelm}. 
The lyrics' vocabulary and syntax are different from poetry and usually more contemporary; therefore, they need to be treated separately. Several works focused on the Rap and Hip Hop genres. They propose to model rhythm and rhyming with unique tokens within the text \cite{xue-etal-2021-deeprapper} or to generate verses conditioned on input keywords and post-process the text to adhere with a rhyme schema \cite{nikolov-etal-2020-rapformer}. 
While most systems are stand-alone, i.e., do not require human intervention, nowadays, we observe a significant demand for human-in-the-loop approaches, that is, models that can help humans better pursue their goals. In this spirit, \citet{ram-etal-2021-saywhat} proposed a songwriter assistant able to consider different aspects of a song, including producing verses with a given metric or that rhyme with a given word.

We join this cause and propose an interactive approach to lyrics generation in English and other 12 languages, which can be conditioned on different song attributes and an arbitrary rhyme schema. Different from \citet{xue-etal-2021-deeprapper}, our approach does not require reversing a text nor implementing architectural changes as in \citet{li-etal-2020-rigid}, allowing us to leverage the knowledge encoded within a PLM easily, yet being able to produce high-quality rhymes as requested by a songwriter. Furthermore, our approach is more flexible than the one proposed in \citet{ram-etal-2021-saywhat} as it allows us to define words each verse should rhyme with or generate a stanza from scratch, given only the desired rhyme schema. Finally, we show that our approach is language-agnostic and propose a unified neural network to produce lyrics in 13 languages.\footnote{Due to the resources all languages are European.}

%%%%%%%%%%%%%%%%%%%%%%%%%%%%%%%%%%%%%%%%%%%%%%%%%%%%%%%%%%%%%%%%%%%%%%
%%%%%%%%%%%%%%%%%%%%%%%%%%%%%%%%%%%%%%%%%%%%%%%%%%%%%%%%%%%%%%%%%%%%%%

\section{Model}
\label{sec:model}
We make use of an encoder-decoder architecture to condition the lyrics generation on a given set of inputs, such as \textit{artist's style}, \textit{title}, \textit{genre}, \textit{topics}, \textit{emotions}, and \textit{rhyme schema}. The input to the model is formatted as follows:

{\small
\medskip
\noindent\texttt{<BOS><title> The River <Artist> Bruce Springsteen <emotions> sad <rhyming\_schema> A B B <EOS>}
\medskip
}

\noindent Here, the model is trained to generate three verses, where the second and third rhyme. However, due to training through conditional language modeling, the last words of the verse tend to attend more heavily on the generated context than the rhyming schema, leading to non-rhyming output.
% as our experiments suggested, more than straightforward fine-tuning is needed to achieve satisfactory results in terms of rhyming. 

% \paragraph{Last Word First (LWF) Strategy} 
We propose the Last Word First (LWF) approach, that relies on anticipating the rhyming word. The last word of a sentence is generated as the first token right after the rhyming symbol and separated by the sentence it belongs to with the special token \texttt{<sep>}, as in the example below (and LWF models in Appendix \ref{sec:lyrics_examples}): 

%This paper proposes the Last Word First (LWF) strategy to train our model, i.e. to additionally prepend the last word of each verse at the beginning of the verse, as follows:  %, which we detail in what follows.

{\small
\medskip
\noindent\texttt{A: river <SEP> We'd go down the river <EOS>}\\
\noindent\texttt{B: dive <SEP> And into the river we'd dive <EOS>}\\
\noindent\texttt{B: ride <SEP> Oh, down the river we'd ride <EOS>}
\medskip
}

\noindent Consult Fig.~\ref{fig:lwf_model}; the green boxes represent the input/output of the model. The lower box is the encoder input that provides information for generating the lyrics. %Instead, 
The left box is the rhyme schema template, passed explicitly to the decoder to enforce that it generates a stance coherent with the input rhyme schema. % at generation time. 
In more detail, the rhyme schema is forced at generation time by detecting when a sentence-end token is generated and forcing the following token to be the next rhyming symbol in the queue. 
% Finally, as one can see, the output differs from what one might typically expect.

With this approach, we can generate coherent sentences that end with rhyming words and quickly identify rhyming patterns since a rhyming word always follows a rhyming symbol. Retaining pretraining knowledge is a key advantage, as opposed to training from scratch. 
% The results in Table \ref{tab:results} confirm this hypothesis, as we produced meaningful text while achieving similar results in terms of rhyming.
%Previous work on reversed language models, i.e., models trained on texts with sentences reversed \cite{xue-etal-2021-deeprapper, li-etal-2020-rigid} were trained from scratch, thus spoiling the knowledge of the pre-trained model and requiring more significant amounts of data. In what follows, we will denote this technique as $RTL$ (Right-To-Left). 
Based on this strategy, we propose the following two models variants:

\paragraph{Plain Last Word First (\textsc{LWF})} the encoder-decoder model is fed with a prompt specifying different desiderata such as: \textit{artist's style}, \textit{title}, \textit{genre}, \textit{topic} and \textit{emotions} and trained by minimising the cross-entropy loss at the token level. As usual for generation models, we use teacher forcing at training time, i.e., to predict the $i$-th token, we feed into the decoder the gold tokens up to time-step $i-1$.
Formally, for each input, we minimise the following loss:
\begin{align}
\label{eq:crossentropy}
    \mathcal{L} &= \frac{1}{N} \sum_i^N \sum_j^{|V|} p_i^j log \; \hat{y}_i^j\\
    \hat{y}_i &= \mathcal{M}(X, t_1, \dots, t_{i-1})
\end{align}
\noindent where $X$ is the input to our model $\mathcal{M}$, $V$ is the model's vocabulary, $t_1, \dots t_{i-1}$ are the gold tokens for the first $i-1$ timesteps, $\mathcal{M}(\dots)$ outputs a vector of logits of size $|V|$ and $\cdot\;^j$ selects the $j$-th element of a vector.

% \begin{figure}[t]
% \centering
% \includegraphics[scale=0.4]{images/model_diagram-plus.png}
% \caption{Drawing of the multitask model.}
% \label{fig:lwf_model_multitask}
% \end{figure}

\begin{table}[t]
    \centering
    % \resizebox{\textwidth}{!}{
    \begin{tabular}{lrrr}
    \toprule
    & Train & Dev & Test \\
    \midrule
    \# Examples & 564K & 3.5K & 3.5K \\
    With Genres & 146K & 587 & 804 \\
    With Emotions & 27K & 62 & 104 \\
    With Topics & 170K & 719 & 893 \\
    \midrule
    Avg. Tokens & 46.92 & 74.90 & 70.79 \\
    Avg. Sentences & 5.84 & 9.77 &9.24 \\
    Avg. Sentence Length & 8.03 & 7.67 & 7.67 \\
    \bottomrule
    \end{tabular}
    % }
    \caption{Statistics for Train, Dev and Test splits of the Genius.com dataset.}
    \label{tab:genius_stats}
\end{table}
\begin{table}[t]
    \centering
    % \resizebox{\textwidth}{!}{
    \begin{tabular}{lrrr}
        \toprule

    & Train & Dev & Test \\
    \toprule
    \# Examples & 2.6M & 10K & 10K\\
    With Genres & 1.2M & 4.3K & 4.5K\\
    % With Emotions & 0 & 0 & 0\\
    With Topics & 170K & 493 & 493 \\
    Languages & 13 & 13 & 13\\
    \midrule
    Avg. Tokens & 40.19  & 39.80 & 39.80 \\
    Avg. Sentences & 6.99 & 6.79 & 6.60  \\
    Avg. Sentence Length & 5.75 & 5.87 &  6.03\\
    \bottomrule
    \end{tabular}
    % }
    \caption{Statistics for Train, Dev and Test splits of the multilingual dataset.}
    \label{tab:multilingual_stats}
\end{table}

\paragraph{Last Word First + Ending Phonetic Representation (\textsc{LWF+EPR})}
Beyond lyrics generation, as the plain \textsc{LWF} model, this variant includes the secondary objective of generating the ending phonetic representation of a word given as input. 
Intuitively, this objective helps inject the word's phonetic features into the model, thus helping to produce more accurate rhymes (see violet boxes in Figure~\ref{fig:lwf_model}). The model is trained through multitasking, by alternating batches between tasks, computing the cross-entropy losses, and updating the model weights separately. We computed the phonetic representation of a word by using CMU Pronouncing Dictionary \cite{weide-etal-1998-cmu}, and, specifically, the \textit{pronouncing} python library.\footnote{\url{https://github.com/aparrish/pronouncingpy}} 
At training time, we use the list of the last words in the lyrics dataset and sample them according to their frequency.

\section{Datasets}
This section details the sources and procedures to recreate the datasets used within this work.
\label{sec:datasets}
\subsection{English Dataset}
For the English data, we selected the top \num{1000} artists according to Spotify\footnote{\url{https://chartmasters.org/most-streamed-artists-ever-on-spotify/}} and downloaded all their songs' lyrics available at \url{https://genius.com}.\footnote{we used python API available at \url{https://lyricsgenius.readthedocs.io}}. We ensure that the language is English \footnote{language detection: \url{https://pypi.org/project/phonemizer/}, \url{https://pypi.org/project/spacy-langdetect/ https://pypi.org/project/stanza/} and spacy's language detector.}. Genius.com offers well-polished lyrics comprising annotations for choruses, pre-choruses, verses, etc. (Table \ref{tab:the_river_full}). 
Since our goal is not to generate the full song lyrics all at once but to create verses that follow a specific rhyme schema and other desiderata, we need to reshape song lyrics data. To this end, we split each song into paragraphs corresponding to different parts, i.e., choruses, bridges, verses, etc. Thus, each item in our dataset is a song paragraph with its song's metadata, i.e., title, artist, genre, topics and emotion (whenever available). Furthermore, to allow a model to generate a stanza based on previous verses, we add to the metadata information the verses of the stanza preceding them, when appropriate.
% for all those sections that are not the first ones.

% In this work, we focus on cross-sentence rhyming, i.e., rhymes that occur between the last word of different sentences within the same paragraph, since this is the most common kind of rhymes in Pop music, as opposed to other rhetorical figures as alliteration.
We focus on rhyming between the last word of different sentences in the same paragraph, as it's the most common type of rhymes in Pop music, as opposed to other rhetorical figures as alliteration.
As Genius.com does not provide the rhyming schema, we computed it for each item, by first tokenising its lyrics. We added the phonetic representation of the last token of each verse\footnote{We used the \textit{phonemizer} python library available at \url{https://github.com/bootphon/phonemizer}}. Then, we compared them pairwise by applying \citet{ghazvininejad-etal-2016-generating}'s algorithm for rhymes and near rhymes in English to assign the same rhyming letter to all words that rhyme together.
For example, given the lyrics in Table \ref{tab:the_river_full}, we assign to the chorus this rhyme schema: \texttt{ABB}.

% For example, given the lyrics in Table \ref{tab:wish_you_were_here}, we assign this rhyme schema: \texttt{AABCDDA}.\footnote{Note that we consider that two identical words rhyme together since they still serve the purpose of rhyming in lyrics, i.e., that of creating a familiar and harmonic sound and pattern in words.}
% \input{tables/wish_you_were_here_verses}

Once the dataset is created, we split it into three subsets for training, development and testing, respectively;
we report their statistics in Table \ref{tab:genius_stats}.

\subsection{Multilingual Dataset}
For languages other than English, we resort to data available within \wasabi\cite{buffa-etal-2021-wasabi}, an extensive database of songs containing lyrics and other metadata about roughly 2M of songs in 21 languages. 
To build our multilingual dataset, we kept pieces in all languages for which we can extract the phonemes\footnote{Please, refer to \url{https://github.com/espeak-ng/espeak-ng/blob/master/docs/languages.md} for the list of \textit{phonemizer} library's supported languages.} and filtered out those languages with less than \num{3000} songs. 
As a result, our dataset covers \num{12} languages plus English. Once we selected the languages, we built the dataset similarly to the English case. However, since \wasabi data is noisier than Genius.com ones, it is not always the case that a song can be clearly divided into sections. Therefore, in all those cases where such splitting is not explicit, we apply a simple heuristic and divide the songs into groups of \num{6} sentences.\footnote{We decided to use \num{6} since that is the average number of sentence for each paragraph in the Genius.com English training set.}
In this case, the rhyme schema is also automatically induced by slightly modifying the algorithm of near rhymes used for English\footnote{For tokenisation, we used \textit{stanza} python library.} by defining the set of vowels for each language of interest.\footnote{We acknowledge that each language may have peculiarities to form rhymes. However, investigating all of them is out of the scope of this work, and it is left as a possible future direction.} 

The final dataset is created by merging all language-specific datasets and splitting them into %three subsets for
training, development and testing subsets; we report the multilingual dataset statistics in Table~\ref{tab:multilingual_stats}.
%, while statistics for language-specific datasets can be found in Table XX of the Appendix.

\section{Experimental Setup}
\label{sec:exp_setup}
This section introduces the research questions we aim to answer throughout our experiments, the results attained, and a human analysis of the lyrics we produced.

\paragraph{Models and training} We carried out our experiments with the T5 \cite{raffel-etal-2020-t5} encoder-decoder architecture for English experiments, through the transformers library.\footnote{\url{https://huggingface.co/docs/transformers/index}} For each one of the models, we used the two decoding strategies described in \ref{sec:results}. we fine-tuned Multilingual T5 \cite[MT5]{xue-etal-2021-mt5} on our multilingual datasets.%, evaluated its ability to rhyme across languages, and automatically evaluated its fluency.

\paragraph{Notation} %For the metrics, we follow the notation just above in section \ref{sec:eval_metrics}. 
We indicate that a model finetunes a pretrained model with *$^\text{Pretrain}$ and with *$^\text{Rand}$ when training from scratch. The sub-incidences indicate the fine-tuning technique, with %*$_\textsc{rtl}$, 
*$_\textsc{lwf}$, and $_\textsc{lwf+epr}$ meaning %Right-To-Left, 
Last-Word-First, and Last-Word-First plus Ending-Phonetic-Representation respectively. Regarding the decoding techniques, BS stands for Beam Search\footnote{We use beam equal $4$} while S+R stands for sampling sentences $k$\footnote{We use $k=20$ in our experiments.} and reranking them according to adherence with the rhyme schema. Examples selected randomly can be found in Appendix \ref{sec:lyrics_examples}.

% \subsection{Models and Training}
% \label{sec:exp_setup.models}

% We also add a plain right-to-left ($\text{rtl}$) model as a reference for a similar technique, \tflargertl.

% Since already with the base instead \tflarge we got better marks on rhyming and worse on Mauve. 

% \subsection{Datasets}
% We carried out experiments with the datasets previously presented in Section \ref{sec:datasets}. According to the multitask training, besides the lyrics dataset, we generate, at training time, a dataset of word-pronounce pairs. We use the list of the last words in the lyrics dataset and sample them according to their frequency.

\subsection{Evaluation Metrics}
\label{sec:eval_metrics}
We evaluate the models with different metrics to provide information on various aspects.

\paragraph{Coherence metrics}
To assess to what extent the model was able to learn the language of lyrics and its structure, we chose two metrics: perplexity (PPL) as a base measure and Mauve \cite{Mauve} for a deeper comparison of the distributions. 
% For a deeper lookup of the second one, check the paper \cite{Mauve}.

We use perplexity as a base measure to assess to what extent the model learnt the language of lyrics. For each song $s$, we consider the model perplexity as follows:
\begin{align}
    &PP(s)=e^{H(s)}\\
    &H(s)=-\sum_{i}^{N}{p(y_i|y_{<i},x)\;log\;p(y_i|y_{<i},x)}
\end{align}
\noindent where, $N$ is the number of tokens in $s$, $y_i$ is the $i$-th token, $y_{<i}$ is the sequence of tokens before $i$ and $x$ is the input data, i.e., artist, title, topics, rhyme schema (as explained in Section \ref{sec:model}).

\paragraph{Rhyming metrics}
% As for evaluating rhyming, we measure the model macro precision concerning the required schema and the false positive rate between tokens that are not supposed to rhyme. Finally, we estimate the ability of the model to generate the number of sentences required by the input and the coverage in terms of necessary rhyming tokens.
To evaluate rhyming, we measure the model's macro precision and false positive rate between tokens that are not supposed to rhyme. We also estimate its ability to generate the required number of sentences and the coverage in terms of necessary rhyming tokens.
Formally, for each song, we compute the Rhyming Precision (RP) and the Rhyming False Positive Rate (R. FP) as follows:
\begin{align*}
    P &= \frac{1}{|R|}\sum_{t_i, t_j}^R \text{rhyme}(t_i, t_j) \\
    FPR &=  \frac{1}{|NR|} \sum_{t_i, t_j}^{NR} 1- \text{rhyme}(t_i, t_j) \\
    % SC &= \frac{1}{|S|} \sum_i^{|S|} \mathds{1}(r_i, s_i)
\end{align*}
\noindent where $P$ is the rhyming precision, i.e., for each pair $(t_i, t_j)$ in the set $R$ of generated tokens that are supposed to rhyme according to the input schema, $\text{rhyme}(t_i, t_j)$\footnote{To evaluate whether two tokens rhyme, we apply the same approach described in Section \ref{sec:datasets}.} evaluates to $1$ if $t_i$ and $t_j$ rhyme and $0$ otherwise. Instead, $FPR$ (False Positive Rate) measures the ratio of token pairs in $NR$, i.e., the set of generated tokens that are not supposed to rhyme while rhyming.
%Finally, $SC$ considers only the starting of each sentence and checks whether the generated text covers all letters in the rhyme schema in the same order. For each correctly covered letter, we add $1$.

\paragraph{Diversity metrics}
To measure the diversity of the generated text, we use distinct metrics. We count the number of $N$-grams that are unique and divide it by the total number of $N$-grams in the text. In the tables 
\ref{tab:results_pretrain} and \ref{tab:results_scratch} 
we can find the values for $N$ equal to $2$, $3$ and $4$ which we denote as $D$-$2$, $D$-$3$, and $D$-$4$, respectively. 

\paragraph{CopyRight metric \copyright}
We made a string matching measure, \copyright, to detect lyrics with original dataset sentences. For each generated output, we compare it to each entry in the entire data set and find the longest subsequence, allowing a wrong token in the middle. If the length of the longest subsequence is above a pre-determined threshold, we choose $20$ for our dataset and tokeniser, we consider the generated output to be at risk of being deemed plagiarised. We also calculate the percentage of this subsequence's length to the generated output's size as references for checking. The final score is the number of generated outputs at risk divided by the number of outcomes. 
% As this is an iterative process over the entire set of generated output and the whole dataset; we use a sentence-transformers model to search the dataset and find the closest 100 entries to the generated output before performing the above comparisons and calculations.

\subsection{Research Questions}
Through our experiments, we aim to answer the following research questions:
\begin{enumerate}
    \item \textbf{Q1:} What is the impact of the \textsc{LWF} strategy in terms of rhyming accuracy?
    \item \textbf{Q2:} Considering that generating lyrics differs from generating standard text, is the knowledge contained in a PLM still relevant regarding rhyming accuracy and text fluency? 
    \item \textbf{Q3:} How does \textsc{LWF} compare against previous SOTA approaches like \textsc{RTL}, right-to-left manner? 
    \item \textbf{Q4:} Since syllables may be relevant when dealing with rhymes, what is the impact of tokenisation on the overall performance?
    \item \textbf{Q5:} Does the phonetic information introduced by multitasking \textsc{LWF+EPR} result in any improvement in terms of rhyming?
    \item \textbf{Q6:} Can we learn a model shared across languages while preserving rhyming accuracy?
\end{enumerate}

\noindent We present models and corresponding results for each of these questions in the next section.

\subsection{Results}
\label{sec:results}
\begin{table*}[t]
    \centering
    \begin{tabular}{lllllccccccc}
        \toprule

    Model & Token. & Decod. & PPL $\downarrow$ & R. P $\uparrow$ & R. FP $\downarrow$ & Mauve $\uparrow$ & \copyright & D-2 & D-3 & D-4 \\
    \toprule
    \tflargeplane & T5 & BS & \textbf{3.50} & 11.36 & 2.31  & 0.0110  & 31.50 & 16.05 & 34.65 & 51.07\\ % &  02.89
    \tflargeplane & T5 & S+R & \textbf{3.50} & 35.44 & 3.68 & 0.0087 & 8.37 & \textbf{25.1} & \textbf{61.07} & 85.50 \\ %  & 3.15
    \midrule
    \midrule
    % \tflargertl & T5 & BS & - & - & - & - & - & - & -  & -\\ % 
    % \tflargertl & T5 & S+R & - & - & - & - & - & - & -  & - \\ % 
    % \midrule
    % \toprule
    \tfbaserandom & T5 & BS & 5.91 & 55.18 & 19.10 & 0.0120 & 32.35 & 7.61  & 25.73 & 50.38 \\ %  & 0.57
    \tfbaserandom & T5 & S+R & 5.91 & \textbf{94.13} & \textbf{11.28} & 0.0152 & \textbf{5.13} & 19.47 & 58.31 & \textbf{87.78} \\ %  & 1.18 
    \midrule
    \tflargepretrain & T5 & BS & 5.88 & 55.82 & 18.99 & 0.0238 & 26.09 & 18.91 & 41.76 & 60.51 \\ % & 2.25
    \tflargepretrain & T5 & S+R & 5.88 & 89.79 & 11.68 & \textbf{0.0240} & 8.52 & 22.36 & 59.66 & 87.32 \\ %  & 1.85
    \midrule 
    \end{tabular}
    % }
    \caption{Results of various models trained on the English dataset. We follow the notation from \ref{sec:eval_metrics} for the metrics and \ref{sec:results} for models and decoding.
    T5-B indicates the base model of T5, T5-L the large version, *$_{\text{LWF}}$ indicates that the model has been trained with last-word-first, while *$_\text{LWF+EPR}$ that the model has been trained on lyrics generation and phoneme generation tasks.}
    \label{tab:results_pretrain}
    \vspace{5mm}
\end{table*}

    % Model & Tokens & Decoder & PPL $\downarrow$ & Rhyme P $\uparrow$ & Rhyme FP $\downarrow$ & Mauve & CopyRight \\
    % \toprule
    % \tflargeplane & T5 & S+R & 3.50 & 35.44 & 0.04 & \hrule & \hrule \\
    % \tflargerandom & T5 & S+R & \hrule & \hrule & \hrule & \hrule & \hrule \\
    % \tflargertl & T5 & S+R & \hrule & \hrule & \hrule & \hrule & \hrule \\
    % \midrule
    % \tflargepretrain & T5 & BS & 5.88 & 55.21 & 15.07 & \hrule & \hrule\\
    % \tflargepretrain & T5 & S+R & 5.88 & 89.79 & 11.68 & \hrule & \hrule \\
    % \midrule 
    % \tflwfplus & T5 & BS & 5.54 & 36.43  & 5.68 & \hrule & \hrule \\
    % \tflwfplus & T5 & S+R & 5.54 & 48.45 & 7.55 & \hrule & \hrule \\

\begin{table*}[t]
    \centering
    \begin{tabular}{lllllccccccc}
        \toprule

    Model & Token. & Decod. & PPL $\downarrow$ & R.P $\uparrow$ & R.FP $\downarrow$ & Mauve $\uparrow$ & \copyright & D-2 & D-3 & D-4 \\
    \toprule
    \tfbaserandom & Word & BS  & 6.79 & 15.38 & \textbf{7.96}  & 0.0046 & 42.16 & 1.87  & 5.89  & 10.79 \\
    \tfbaserandom & Word & S+R & 6.79 & 59.40 & 8.71  & 0.0056 & \textbf{5.11}  & 8.20  & 33.65 & 58.43 \\ 
    \midrule
    \tfbaserandom & T5   & BS  & \textbf{5.91} & 55.18 & 19.10 & 0.0120 & 32.35 & 7.61  & 25.73 & 50.38 \\ %  & 0.57
    \tfbaserandom & T5   & S+R & \textbf{5.91} & \textbf{94.13} & 11.28 & \textbf{0.0152} & 5.13  & \textbf{19.47} & \textbf{58.31} & \textbf{87.78} \\ %  & 1.18  
    \bottomrule
    \end{tabular}
    % }
    \caption{Tokenizer comparison: Results of two models trained on the English dataset from scratch, one with a word tokenizer (Word) and the other with the default tokenizer (T5), as indicated in the column Token. We follow the notation from \ref{sec:eval_metrics} for the metrics and \ref{sec:results} for models and decoding.}
    \label{tab:results_scratch}
    \vspace{4mm}
\end{table*}
\begin{table}[t]
    \centering
    % \resizebox{\textwidth}{!}{
    \begin{tabular}{llccc}
        \toprule
    Model & Decod. & R.P $\uparrow$ & R.FP $\downarrow$ \\ % PPL $\downarrow$ &
    \toprule
    \tflargepretrain & BS & 55.82 & 18.99 \\ % 5.88 &
    \tflargepretrain & S+R & \textbf{89.79} & 11.68\\ % 5.88 &
    \midrule 
    \tflwfplus & BS & 36.43  & \textbf{5.68} \\ %  \textbf{5.54} &
    \tflwfplus & S+R & 48.45 & 7.55 \\ %   \textbf{5.54} &
    \midrule 
    \tflargertlpre & BS & 39.3  & 6.76 \\ % ???
    \tflargertlpre & S+R & 52.5 & 6.85 \\ % 
    \bottomrule
    \end{tabular}
    \caption{Results of the comparison between Last Word First \tflargepretrain, the Last Word First + Ending Phonetic Representation \tflwfplus and Right-to-Left \tflargertlpre. We follow the notation from \ref{sec:eval_metrics} for the metrics and \ref{sec:results} for models and decoding.}
    \label{tab:results_RTL_EPR}
\end{table}

\begin{table}[t]
    \centering
    % \resizebox{\textwidth}{!}{
    \begin{tabular}{llrrr}
        \toprule

    Language & R.P & R.FP & Support\\
    \midrule
    English & \textbf{54.78} & 9.96 & 1593\\
    French & 52.2 & 15.53 & 1355\\
    Dutch & 46.83 & 11.89 & 258\\
    German & 42.34 & 8.87 & 1386\\ \hdashline
    Danish & 39.25 & 6.49 & 53\\
    Swedish & 30.27 & 8.09 & 197\\
    Norwegian & 27.44 & 8.80 & 88\\ \hdashline
    Portuguese & 36.6 & 11.64 & 841\\
    Italian & 35.82 & 7.91 & 742\\
    Spanish & 32.62 & 8.90 & 2485\\ \hdashline
    Croatian & 35.61 & 10.81 & 90\\
    Polish & 34.94 & 9.23 & 338\\ \hdashline
    Finnish & 24.09 & \textbf{8.09} & 370\\
    \midrule
    Micro AVG & 40.99 & 10.19 & \\
    \bottomrule
    \end{tabular}
    % }
    \caption{Results of the multilingual model by fine-tuning mT5 with LWF technique. We used nucleus with 0.92 for top p as a sampling strategy}
    \label{tab:results_multilingual}
\end{table}

% \begin{table}[t]
%     \centering
%     % \resizebox{\textwidth}{!}{
%     \begin{tabular}{llrrr}
%         \toprule

%     Language & Rhyme P & Rhyme FP & Support\\
%     \midrule
%     Croatian & 32.83 & 9.26 & 90\\
%     Danish & 37.62 & 7.56 & 53\\
%     Dutch & 48.83 & 13.99 & 258\\
%     English & 52.06 & 9.88 & 1593\\
%     Finnish & 25.29 & 7.44 & 370\\
%     French & 49.17 & 14.12 & 1355\\
%     German & 36.10 & 7.19 & 1386\\
%     Italian & 33.69 & 7.23 & 742\\
%     Norwegian & 27.67 & 8.40 & 88\\
%     Polish & 31.66 & 9.04 & 338\\
%     Portuguese & 35.11 & 10.58 & 841\\
%     Spanish & 30.49 & 8.10 & 2485\\
%     Swedish & 27.77 & 6.99 & 197\\
%     % % Slovak & 31.16 & 5.94 & 101\\
%     % % Turkish & 39.64 & 13.15 & 103\\
%     \midrule
%     \midrule
%     Micro AVG & 40.99 & 10.19 & \\
%     \bottomrule
%     \end{tabular}
%     % }
%     \caption{Results of the multilingual model by fine tuning mT5 with LWF technique. We used a sampling strategy}
%     \label{tab:results_multilingual}
% \end{table}

\paragraph{English Evaluation} Main results are in Table \ref{tab:results_pretrain}. To answer \textbf{Q1}, we fine-tuned models with the LWF strategy (e.g. \tflarge) and compared them against the same architecture trained on plain data, i.e., left-to-right without LWF (e.g. \tflargeplane). 
%The first model corresponds to previous existing approaches, \tflargeplane ~ is trained in a plain manner, i.e., without prepending to each sentence its last word (Section \ref{sec:exp_setup.models}).
The \textsc{LWF} models (Section \ref{sec:model}), based on the % simple yet effective 
approach proposed in this paper, attain consistently better results in terms of RP, R.FP, and Mauve. The best system in terms of rhyming is instead \tfbase ~ with Random initialisation and S+R, also beating its larger and pretrained counterpart (\tflarge). Though, as Mauve indicates and we show in Table \ref{tab:aivshuman}, \tfbase ~ produces much less meaningful lyrics, not creating human-like songs. 
The decoding technique highly affects the rhyming performance and copyright. While BS led to modest performance, we could boost performance with S+R in both aspects. As expected, Mauve and the three diversity metrics get better results with a sampling decoding.

To investigate whether the knowledge in a PLM is relevant to lyrics generation (\textbf{Q2}), we provide results attained by a T5-base model trained from scratch (\tfbaserandom) and compare against a pretrained version (\tflargepretrain). 
After comparing \tflargepretrain to \tflargerandom, we chose the base model for better results.
Data input during fine-tuning versus model size may cause the observed variation.
% In preliminary experiments, we compared \tflargepretrain against \tflargerandom, but finally opted for the base model as it produced better results. 
% We assume this is due to the amount of data introduced during finetuning compared to the size of the model. 
\tflargeplane ~ attains the best score on perplexity across the board, yet its RP and R.FP are the worst. On the other hand, \tflargepretrain obtains the best results in Mauve, a coherence metric with a higher correlation with humans. The superior coherence of the pretrained models will be further confirmed in Section~\ref{humavsai}, where we present human evaluation.

For \textbf{Q3}, we compare \tflarge ~ against its counterpart \tflargertl in Table \ref{tab:results_RTL_EPR}, showing that although this is an improvement over \tflargepretrain, it is still far from the results of our \textsc{LWF} method.

For \textbf{Q4}, we trained \tfbase with two different tokenisers, the original one for T5-base and a word-level one. In Table \ref{tab:results_scratch}, we show that the word tokeniser, which may not split the end of the words into tokens, produces worse results even when training from scratch.

%We clarify any model that includes phonetic knowledge through the multitasking approach improves the rhyming (Q5) by adding an $\text{EPR}$ notation, e.g. \tflwfplus.
For \textbf{Q5}, we observe in Table \ref{tab:results_RTL_EPR} that \textsc{EPR} does not affect the model positively. While outperforming \tflargeplane, \tflwfplus ~ attains worse scores than both \tfbase ~ and \tflarge, suggesting that generating phonemes does not inject proper knowledge to ease the rhyming generation process. 

In addition, although prompt conditioning has been previously studied, Section \ref{sec:conditional} of the appendix shows that, even if these conditions are not explicitly stated in the objective function, they correlate well with human-generated lyrics.

\paragraph{Multilingual Evaluation}
To address \textbf{Q6}, in Table \ref{tab:results_multilingual} we report the results breakdown of the multilingual model in each language. Results indicate that learning rhyming across languages is quite complicated. Some languages are more complex, with Finnish, Norwegian, and Swedish having the worst scores and English, French, and Dutch having the best. This is mainly due to the nature of the pretraining data of mT5 \cite{xue-etal-2021-mt5}, where most text is in English followed, at a considerable distance but a similar amount among them, by Spanish, German, and French. 
Less frequently represented languages in mT5 and our dataset (see Table~ \ref{tab:multilingual_datasetbylanguage} in the Appendix) correspond to languages with lower scores. 

We can also observe shared learning between languages from the same phonetic family. While Spanish is the most common in our dataset, the results are lower than in French or German, with phonetically Latin languages a lower score in general. The case of Finish is more remarkable since, not being Indo-European, it has no other supporting languages, giving the worst result. Even Danish and Croatian have better metrics with a lower representation in both datasets.

Worse results are expected when converting a model into a multilingual, especially in our case, where the dataset is significantly smaller. On average, model performance is poor (\num{40.99}), more than \num{40} points lower in terms of Rhyme Precision than the English-only model. While the model proved capable (to some extent) of deriving rhyming rules in English with text data only, it fails to do so when presented with data in several languages while showing some correlation based on phonetics. Indeed, rhyming is strictly tight to the way words are pronounced. 
% Therefore, phonetic features are essential for multilingual models learning to rhyme. 
% TODO: the rest of the paragraph could be skipt if needed.
Recently, phonetic representation of text has been proposed as interlingual \cite{leong-whitenack-2022-phone} with encouraging results, and, in future work, it would be interesting to explore this idea also in the context of lyrics generation. Examples can be found in Appendix \ref{sec:multilingual_examples}

%The best language is English, as usual, mainly due to the nature of the pretraining data of Multilingual T5, where most text is in \cite{xue-etal-2021-mt5}. 
%This is further corroborated by the fact that the most frequent language in our multilingual dataset is Spanish, as the support shows, in which our model attains unsatisfactory results.
%This poor performance is because the model has access to text form only, thus making it hard to learn rhyming patterns across languages.

\section{Human Evaluation}
\label{humavsai}
\begin{table}[t!]
    \centering
    \resizebox{\columnwidth}{!}{
    \begin{tabular}{lrrr}
    \toprule
    & Human & \tflargepretrain & \tfbaserandom \\
    \midrule
    Correctness & $2.61 \pm 0.07$ & $\textbf{2.41} \pm 0.03$ & $2.13 \pm 0.07$ \\
    \midrule
    Meaningfulness & $2.43 \pm 0.02$ & $\textbf{2.19} \pm 0.02$ & $1.69 \pm 0.06$ \\
    \midrule
    Is-Human Rate & $79.67 \pm 2.43$ & $\textbf{57.00} \pm  1.81$ & $23.43 \pm 1.39$ \\
    % \midrule
    % IAA & --- & --- & --- \\
    \bottomrule
    \end{tabular}
    }
    \caption{Results on the human-evaluation tasks. Correctness: 3 is maximum, 1 is minimum; Meaningfulness: 3 is maximum, 1 is minimum; Is-Human Rate: rate at which annotators annotated a paragraph from the reference system as human.}
    % Is-AI Rate: the rate at which annotators annotated a paragraph the reference system as AI.}
    \label{tab:aivshuman}
    \vspace{-3mm}
\end{table}

Given the open-domain nature of text generation, automatic evaluation can often be inaccurate. In this section, we detail experiments where human annotators assess the quality of generated lyrics.

\subsection{Setup}
To evaluate lyrics quality, we created an annotation that measures grammatical accuracy, meaningfulness of the text and if it was written by a human.The annotators were three English-speaking university students, not involved in the project.

We sampled 100 snippets from our test set and used their metadata (artist, rhyme schema, etc.) to generate as many texts from the two best models in Table \ref{tab:results_pretrain}, i.e., \tfbaserandom ~ (line \num{4}) and \tflargepretrain ~ (line \num{6}). Hence, for each model, we have \num{200} items (100 paragraphs written by humans and 100 automatically created). We shuffled the items and asked three annotators to review them and assign scores to the following three categories:
\begin{enumerate}
    \item \textbf{Correctness}:
    following \citet{li-etal-2020-rigid}, annotators had to rate lyrics with
    3, grammatically correct; 2, readable but with some grammar mistakes; and 1, unreadable.
    \item \textbf{Meaningfulness}:
    3, meaningful text; 2, the text has some meaning but is expressed confusingly; and 1, the text has no meaning.
    \item \textbf{Is it from a Human?}: annotators were asked whether the presented text was written by a human or automatically.
\end{enumerate}
\subsection{Results}
In Table \ref{tab:aivshuman}, we report the results of our human evaluation. As previously stated, \tflarge ~ attains results in terms of correctness (\num{2.41}) and meaningfulness (\num{2.19}) close to that assigned to lyrics written by humans, i.e., \num{2.61} and \num{2.43} for correctness and meaningfulness, respectively. On the opposite, \tfbase, while still producing grammatically correct texts, their meaningfulness is much lower than human lyrics. Finally, \tflarge ~ generations are classified as written by humans $57\%$ of the time, and, surprisingly, human products are recognised as such $79\%$ of the times. We believe that this is due to the high percentage of lyrics without formal meaning.

\section{Conclusion}
We presented a new approach to enhancing rhyming in a PLM and a first attempt to scale lyrics generation across languages. Our framework provides a tool for the composer to automatically generate paragraphs given several desiderata, i.e., artist's style, song title, song genre, emotions, topics, and rhyme schema. The proposed method proved more effective than fine-tuning lyrics data and coherent lyrics without too much risk of copyright infringement. It produces meaningful and grammatically correct texts by reassembling human songs almost 6 out of 10 (according to human annotators). Furthermore, its accuracy in following the given rhyme schema is nearly $90\%$.

In future work, we aim to focus on biases that affect our model and approaches to mitigate them. Another area that requires further investigation is multilingualism, where performance still needs to be improved from the English one.

\section{Limitations}
One limitation of all analysed models is the need for more control over the language used by the model. Indeed, when specific genres are requested, e.g., rap and hip hop, the model may produce paragraphs interpreted as racist or insulting to certain minorities (e.g., women). This is a huge issue that has not been addressed systematically in the context of lyrics generation, causing, among other things, issues and concerns outside the scientific community.\footnote{\href{https://www.theguardian.com/music/2022/aug/24/major-record-label-drops-offensive-ai-rapper-after-outcry-over-racial-stereotyping}{https://www.theguardian.com/music/2022/aug/24/major-record-label-drops-offensive-ai-rapper-after-outcry-over-racial-stereotyping}} In this paper we did not address this issue directly but instead proposed a study focused on rhyming and multilingualism. We intend to conduct this research shortly, focusing on mitigating biases and actively controlling the kind of language when generating lyrics.

Another limitation of the proposed approach lies in the algorithm checking whether two words rhyme. While designed for English, we adapted it to work for most European languages. However, each language may have its exceptions, which we might have neglected.

% \clearpage
\bibliography{anthology,custom}
\bibliographystyle{acl_natbib}

\clearpage
\onecolumn

\appendix
\section{Appendix}
\subsection{Examples of Generated Lyrics}
% We provide examples of the data generated using the models in \ref{tab:results_pretrain} and \ref{tab:results_scratch} given two different prompts. We want to acknowledge the presence of slurs and offensive language in certain generated outputs presented in this publication. The language models utilised in this research have the potential to generate text that may include offensive or inappropriate language present in the training data. These outputs were automatically generated and were not intentionally included by the authors. 

Using two randomly selected prompts, we offer illustrations of the outputs generated by the models discussed in section \ref{sec:exp_setup}. We must recognise the existence of slurs and offensive language within some of these generated outputs. It should be noted that the language models employed in this study can produce text that may encompass offensive or inappropriate language originating from the training data. We wish to emphasise that these outputs were automatically generated and were not deliberately included by the authors.

\label{sec:lyrics_examples}

\begin{table}[!htb]
\centering
\begin{tabular}{llp{0.7\linewidth}}
\toprule
\textbf{Model} &
  \multicolumn{1}{c}{\textbf{Tokenizer}} &
  \multicolumn{1}{c}{\textbf{generation}} \\
  
  T5-L &
  T5 &
\begin{tabular}{p{\linewidth}}RHYME\_A\textless{}sentence\_end\textgreater\\      RHYME\_B 'Cause I'm a fool for you\textless{}sentence\_end\textgreater\\      RHYME\_C Look at me now\textless{}sentence\_end\textgreater\\      RHYME\_D Oh, oh\textless{}sentence\_end\textgreater\\      RHYME\_D Ooo, look at me now, now yea heah heaheh, hahaha, yeah\textless{}sentence\_end\textgreater\\      RHYME\_D\textless{}sentence\_end\textgreater{}\end{tabular} \\
\midrule

\tfbaserandom &
  Word &
  \begin{tabular}{p{\linewidth}}RHYME\_A   now\textless{}SEP\textgreater I'm ready now\textless{}sentence\_end\textgreater\\      RHYME\_B down\textless{}SEP\textgreater Don't wanna put me down\textless{}sentence\_end\textgreater\\      RHYME\_C around\textless{}SEP\textgreater 'Cause you're not   around\textless{}sentence\_end\textgreater\\      RHYME\_D me\textless{}SEP\textgreater You should be ready for me\textless{}sentence\_end\textgreater\\      RHYME\_D see\textless{}SEP\textgreater So you can see\textless{}sentence\_end\textgreater\\      RHYME\_D be\textless{}SEP\textgreater What you gotta be\textless{}sentence\_end\textgreater{}\end{tabular} \\
  \midrule

\tfbaserandom &
  T5 &
  \begin{tabular}{p{\linewidth}}RHYME\_A me \textless{}SEP\textgreater I'm me   \textless{}sentence\_end\textgreater\\       RHYME\_B oh \textless{}SEP\textgreater Oh, oh   \textless{}sentence\_end\textgreater\\       RHYME\_C \textless{}unk\textgreater \textless{}SEP\textgreater   \textless{}unk\textgreater \textless{}unk\textgreater \textless{}sentence\_end\textgreater\\       RHYME\_D \textless{}unk\textgreater \textless{}SEP\textgreater   Yeah, yeah - - \textless{}sentence\_end\textgreater\\       RHYME\_D me \textless{}SEP\textgreater And I know   you're all for me ( Yeah ) \textless{}sentence\_end\textgreater\\       RHYME\_D Yeah \textless{}SEP\textgreater ( Oh ) Yeah   \textless{}sentence\_end\textgreater{}\end{tabular} \\
 \midrule
\tflargepretrain &
  T5 &
  \begin{tabular}{p{\linewidth}}RHYME\_A now\textless{}SEP\textgreater Look at me   now\textless{}sentence\_end\textgreater\\      RHYME\_B you\textless{}SEP\textgreater I'm so in love with you\textless{}sentence\_end\textgreater\\      RHYME\_C do\textless{}SEP\textgreater And I don't know what to do\textless{}sentence\_end\textgreater\\      RHYME\_D me\textless{}SEP\textgreater But you've got a hold on me\textless{}sentence\_end\textgreater\\      RHYME\_D see\textless{}SEP\textgreater You're the only one I see\textless{}sentence\_end\textgreater\\      RHYME\_D be\textless{}SEP\textgreater That I wanna be\textless{}sentence\_end\textgreater{}\end{tabular} \\ \bottomrule
\end{tabular}
\caption{"Random examples generated with Beam Search decoding by the models from Tables \ref{tab:results_pretrain} and \ref{tab:results_scratch} for the prompt: "\textless{}title\textgreater Look at Me Now\textless{}artist\textgreater Charlie Puth\textless{}schema\textgreater{}RHYME\_A RHYME\_B RHYME\_C RHYME\_D RHYME\_D RHYME\_D\textless{}/s\textgreater{}".}
\label{tab:bs_examples_v1}
\end{table}

\begin{table}[!htb]
\centering
\begin{tabular}{llp{0.7\linewidth}}
\toprule
\textbf{Model} &
  \multicolumn{1}{c}{\textbf{Tokenizer}} &
  \multicolumn{1}{c}{\textbf{generation}} \\

T5-L &
  T5 &
  \begin{tabular}{p{\linewidth}}RHYME\_A\textless{}sentence\_end\textgreater\\      RHYME\_B\textless{}sentence\_end\textgreater\\      RHYME\_C ooh\textless{}sentence\_end\textgreater\\      RHYME\_D\textless{}sentence\_end\textgreater\\      RHYME\_D....\textless{}sentence\_end\textgreater\\      RHYME\_D..\textless{}sentence\_end\textgreater{}\end{tabular} \\

    \midrule
\tfbaserandom &
  Word &
  \begin{tabular}{p{\linewidth}}RHYME\_A eyes\textless{}SEP\textgreater See in the dark,   I was lookin' past the eyes\textless{}sentence\_end\textgreater\\      RHYME\_B die\textless{}SEP\textgreater I used to lie like I would never   die\textless{}sentence\_end\textgreater\\      RHYME\_C you\textless{}SEP\textgreater And no good for you\textless{}sentence\_end\textgreater\\      RHYME\_D me\textless{}SEP\textgreater Even though I really cared that you would let   me\textless{}sentence\_end\textgreater\\      RHYME\_D see\textless{}SEP\textgreater That's not you, now I see\textless{}sentence\_end\textgreater\\      RHYME\_D be\textless{}SEP\textgreater Just remember how it'd always be\textless{}sentence\_end\textgreater{}\end{tabular} \\
\midrule

\tfbaserandom &
  T5 &
  \begin{tabular}{p{\linewidth}}RHYME\_A shit \textless{}SEP\textgreater I want you been   talking talking shit \textless{}sentence\_end\textgreater\\       RHYME\_B yeah \textless{}SEP\textgreater Oh, oh,   yeah \textless{}sentence\_end\textgreater\\       RHYME\_C this \textless{}SEP\textgreater Well, I can   feel this \textless{}sentence\_end\textgreater\\       RHYME\_D say \textless{}SEP\textgreater Yeah, you,   say \textless{}sentence\_end\textgreater\\       RHYME\_D way \textless{}SEP\textgreater You really   get this way ( It's all ) \textless{}sentence\_end\textgreater\\       RHYME\_D away \textless{}SEP\textgreater But baby,   baby get so hard ( Baby get a yes ) ( Yeah now.. ( Oh ) yeah ) just.   \textless{}sentence\_end\textgreater{}\end{tabular} \\
    \midrule
\tflargepretrain &
  T5 &
  \begin{tabular}{p{\linewidth}}RHYME\_A love\textless{}SEP\textgreater I'm dancin', all   I am is love\textless{}sentence\_end\textgreater\\      RHYME\_B you\textless{}SEP\textgreater When I find you\textless{}sentence\_end\textgreater\\      RHYME\_C me\textless{}SEP\textgreater But that's not enough, now you know   me\textless{}sentence\_end\textgreater\\      RHYME\_D go\textless{}SEP\textgreater Don't wanna let go\textless{}sentence\_end\textgreater\\      RHYME\_D know\textless{}SEP\textgreater 'Cause all that I see is what you need to   know\textless{}sentence\_end\textgreater\\      RHYME\_D so\textless{}SEP\textgreater Look at me now, and see that you love me   so\textless{}sentence\_end\textgreater{}\end{tabular} \\
  \bottomrule
\end{tabular}
\caption{Random examples generated with the  Sampling and
Reranking decoding stratgey by the models from Tables \ref{tab:results_pretrain} and \ref{tab:results_scratch} for the prompt: "\textless{}title\textgreater Look at Me Now\textless{}artist\textgreater Charlie Puth\textless{}schema\textgreater{}RHYME\_A RHYME\_B RHYME\_C RHYME\_D RHYME\_D RHYME\_D\textless{}/s\textgreater{}".}
\label{tab:rs_examples_v1}
\end{table}
%  \begin{table}[ht]
%     \centering
%     \begin{tabular}{p{\columnwidth}}
%         \toprule
%         \text{[Input]}\\
%         <prompt> <title> Why Did You Leave Me Darling <artist> The Temptations 
%         <schema> ABCDEECC </prompt>\\
%         \\
%         \text{[Generation]}\\
%          I'm so glad I found you \\
%          You're the one for me \\
%          Oh, baby \\
%          ( You made a fool out of me, oh baby ) \\
%          Yeah, yeah \\
%          Ooh, now \\
%          Baby ( Baby, Baby \\
%          Can you hear me? Yeah \\
%          </s>\\
%         \bottomrule
%     \end{tabular}
%     \caption{Random example generated by \tfbaserandom with T5 tokenizer}
%     \label{tab:T5_LWF_Large_Pretrained}
% \end{table}

\begin{table}
\centering
\begin{tabular}{llp{0.7\linewidth}}
\toprule

\textbf{Model} &
  \multicolumn{1}{c}{\textbf{Tokenizer}} &
  \multicolumn{1}{c}{\textbf{generation}} \\
  
  T5-L &
  T5 &
\begin{tabular}{p{\linewidth}}
RHYME\_A\textless{}sentence\_end\textgreater{}\\
RHYME\_B\textless{}sentence\_end\textgreater{}\\
RHYME\_C\textless{}sentence\_end\textgreater{}\\
RHYME\_D\textless{}sentence\_end\textgreater{}\\
RHYME\_E\textless{}sentence\_end\textgreater{}\\
RHYME\_F\textless{}sentence\_end\textgreater{}\\
RHYME\_E..... sex laws\textless{}sentence\_end\textgreater{}\\
RHYME\_G.\textless{}sentence\_end\textgreater{}
\end{tabular} \\
\midrule

\tfbaserandom &
  Word &
  \begin{tabular}{p{\linewidth}}
  RHYME\_A   eyes\textless{}SEP\textgreater When you close your eyes\textless{}sentence\_end\textgreater\\      RHYME\_B you\textless{}SEP\textgreater I'm with you\textless{}sentence\_end\textgreater\\      RHYME\_C time\textless{}SEP\textgreater It's been a long time\textless{}sentence\_end\textgreater\\      RHYME\_D mind\textless{}SEP\textgreater But you're on my mind\textless{}sentence\_end\textgreater\\      RHYME\_E me\textless{}SEP\textgreater And you don't know what to do with   me\textless{}sentence\_end\textgreater\\      RHYME\_F do\textless{}SEP\textgreater 'Cause what you do\textless{}sentence\_end\textgreater\\      RHYME\_E be\textless{}SEP\textgreater What you wanna be\textless{}sentence\_end\textgreater\\      RHYME\_G you\textless{}SEP\textgreater So what are you?\textless{}sentence\_end\textgreater{}
  \end{tabular} \\
  \midrule

\tfbaserandom &
  T5 &
  \begin{tabular}{p{\linewidth}}
  RHYME\_A \textless{}unk\textgreater{} \textless{}SEP\textgreater{} \textless{}unk\textgreater{}, \textless{}unk\textgreater{} \textless{}sentence\_end\textgreater{}
 RHYME\_B \textless{}unk\textgreater{} \textless{}SEP\textgreater{} I've got to get to the \textless{}unk\textgreater{} - - \textless{}sentence\_end\textgreater{}
 RHYME\_C \textless{}unk\textgreater{} \textless{}SEP\textgreater{} And I can't get enough of my own own's \textless{}sentence\_end\textgreater{}
 RHYME\_D \textless{}unk\textgreater{} \textless{}SEP\textgreater{} Oh, oh - oh \textless{}sentence\_end\textgreater{}
 RHYME\_E yeah \textless{}SEP\textgreater{} Yeah, yeah \textless{}sentence\_end\textgreater{}
 RHYME\_F \textless{}unk\textgreater{} \textless{}SEP\textgreater{} Ooh - ooh \textless{}sentence\_end\textgreater{}
 RHYME\_E yea \textless{}SEP\textgreater{} Hey, yea \textless{}sentence\_end\textgreater{}
 RHYME\_G \textless{}unk\textgreater{} \textless{}SEP\textgreater{} Ah, ah - ah \textless{}sentence\_end\textgreater{}
  \end{tabular} \\
 \midrule
\tflargepretrain &
  T5 &
  \begin{tabular}{p{\linewidth}}
  RHYME\_A law\textless{}SEP\textgreater{} Sex laws, sex law\textless{}sentence\_end\textgreater{}
RHYME\_B laws\textless{}SEP\textgreater{} There's no escape from the sexx laws\textless{}sentence\_end\textgreater{}
RHYME\_C man\textless{}SEP\textgreater{} I'm just a man\textless{}sentence\_end\textgreater{}
RHYME\_D law\textless{}SEP\textgreater{} Sex laws..\textless{}sentence\_end\textgreater{}
RHYME\_E me\textless{}SEP\textgreater{} You can't take it from me\textless{}sentence\_end\textgreater{}
RHYME\_F you\textless{}SEP\textgreater{} 'Cause if you\textless{}sentence\_end\textgreater{}
RHYME\_E be\textless{}SEP\textgreater{} Then you're gonna be\textless{}sentence\_end\textgreater{}
RHYME\_G law\textless{}SEP\textgreater{} A victim of the Sexx law!\textless{}sentence\_end\textgreater{}
  \end{tabular} \\ \bottomrule
\end{tabular}
\caption{Random examples generated with Beam Search decoding by the models from Tables \ref{tab:results_pretrain} and \ref{tab:results_scratch} for the prompt: "\textless{}title\textgreater{} Sexx Laws (Malibu remix)\textless{}artist\textgreater{} Beck\textless{}schema\textgreater{}RHYME\_A RHYME\_B RHYME\_C RHYME\_D RHYME\_E RHYME\_F RHYME\_E RHYME\_G\textless{}/s\textgreater{}". Note: the tokenization of the prompt yeilds unkown tokens, thereby influincing the generation. Throughout the generation process, we did not force any of the models to skip \textless{}unk\textgreater{} tokens. }
\label{tab:bs_examples_v1}
\end{table}

\begin{table}
\centering
\begin{tabular}{llp{0.7\linewidth}}
\toprule
\textbf{Model} &
  \multicolumn{1}{c}{\textbf{Tokenizer}} &
  \multicolumn{1}{c}{\textbf{generation}} \\

T5-L &
  T5 &
  \begin{tabular}{p{\linewidth}}
  RHYME\_A Asked the judge for the cost\textless{}sentence\_end\textgreater{}
RHYME\_B "Doctor can I make a small complaint?"\textless{}sentence\_end\textgreater{}
RHYME\_C And he asked for ten of her dollars\textless{}sentence\_end\textgreater{}
RHYME\_D "What's the point in making me suffer?"\textless{}unk\textgreater{}\textless{}sentence\_end\textgreater{}
RHYME\_E *ahem*\textless{}sentence\_end\textgreater{}
RHYME\_F *yeah\textless{}sentence\_end\textgreater{}
RHYME\_E ********uh-oh!\textless{}sentence\_end\textgreater{}
RHYME\_G.\textless{}sentence\_end\textgreater{}
  \end{tabular} \\
    \midrule
\tfbaserandom &
  Word &
  \begin{tabular}{p{\linewidth}}
  RHYME\_A live\textless{}SEP\textgreater{} I feel in my soul, I'm gonna live\textless{}sentence\_end\textgreater{}
RHYME\_B rain\textless{}SEP\textgreater{} Flying round and around in the rain\textless{}sentence\_end\textgreater{}
RHYME\_C away\textless{}SEP\textgreater{} The only thing I know when my troubles fly away\textless{}sentence\_end\textgreater{}
RHYME\_D blue\textless{}SEP\textgreater{} In a red light, there's the sky like the blue\textless{}sentence\_end\textgreater{}
RHYME\_E me\textless{}SEP\textgreater{} So don't throw out the oceans inside me\textless{}sentence\_end\textgreater{}
RHYME\_F feel\textless{}SEP\textgreater{} Don'Cause I can've got the darkest thing about how I want the way I to feel\textless{}sentence\_end\textgreater{}
RHYME\_E free\textless{}SEP\textgreater{} It'll be over myself, why are we in front row free?\textless{}sentence\_end\textgreater{}
RHYME\_G say\textless{}SEP\textgreater{} There're no quiet life and there is nothing left to say\textless{}sentence\_end\textgreater{}
  \end{tabular} \\
\midrule

\tfbaserandom &
  T5 &
  \begin{tabular}{p{\linewidth}}
  RHYME\_A room \textless{}SEP\textgreater{} The \textless{}unk\textgreater{} \textless{}unk\textgreater{} has \textless{}unk\textgreater{} into a room \textless{}sentence\_end\textgreater{}
 RHYME\_B again \textless{}SEP\textgreater{} I don't ever ever again \textless{}sentence\_end\textgreater{}
 RHYME\_C face \textless{}SEP\textgreater{} My soul is in my face \textless{}sentence\_end\textgreater{}
 RHYME\_D you \textless{}SEP\textgreater{} Well I never ever wanted time to let you \textless{}sentence\_end\textgreater{}
 RHYME\_E know \textless{}SEP\textgreater{} And I'll be back when I say I didn'didn didn oh so you thought I knew I would know \textless{}sentence\_end\textgreater{}
 RHYME\_F \textless{}unk\textgreater{} \textless{}SEP\textgreater{} For the second that I had \textless{}unk\textgreater{} \textless{}sentence\_end\textgreater{}
 RHYME\_E no \textless{}SEP\textgreater{} That I, I was gonna get to get, get a get from a better for a \textless{}unk\textgreater{} with a chance, a a no \textless{}sentence\_end\textgreater{}
 RHYME\_G away \textless{}SEP\textgreater{} Oh, Oh! \textless{}sentence\_end\textgreater{}
  \end{tabular} \\
    \midrule
\tflargepretrain &
  T5 &
  \begin{tabular}{p{\linewidth}}
  RHYME\_A you\textless{}SEP\textgreater{} It's like being caught out and not supposed to love you\textless{}sentence\_end\textgreater{}
RHYME\_B it\textless{}SEP\textgreater{} And it'll be hard to say you're not in it\textless{}sentence\_end\textgreater{}
RHYME\_C baby\textless{}SEP\textgreater{} But in my eyes, baby\textless{}sentence\_end\textgreater{}
RHYME\_D truth\textless{}SEP\textgreater{} There'd be no use in escaping the truth\textless{}sentence\_end\textgreater{}
RHYME\_E me\textless{}SEP\textgreater{} Just go with me, go along with I\textless{}sentence\_end\textgreater{}
RHYME\_F man\textless{}SEP\textgreater{} Until the day that we met, man,,man\textless{}sentence\_end\textgreater{}
RHYME\_E see\textless{}SEP\textgreater{} I will never forget the world I see\textless{}sentence\_end\textgreater{}
RHYME\_G yeah\textless{}SEP\textgreater{} In my sight and in your mouth, yeah\textless{}sentence\_end\textgreater{}
  \end{tabular} \\
  \bottomrule
\end{tabular}
\caption{Random examples generated with the  Sampling and
Reranking decoding stratgey by the models from Tables \ref{tab:results_pretrain} and \ref{tab:results_scratch} for the prompt: "\textless{}title\textgreater{} Sexx Laws (Malibu remix)\textless{}artist\textgreater{} Beck\textless{}schema\textgreater{}RHYME\_A RHYME\_B RHYME\_C RHYME\_D RHYME\_E RHYME\_F RHYME\_E RHYME\_G\textless{}/s\textgreater{}".}
\label{tab:rs_examples_v1}
\end{table}

% \subsection{Multilingual Lyrics Examples and Dataset example}
% \section{Multilingual Lyrics Examples and Dataset example}
\label{sec:multilingual_examples}
\begin{table}[h]
\centering
\scalebox{1}{
    \begin{tabular}{p{\columnwidth}}
        \toprule
        \text{[Input]}\\
        <title> Você Não Sabe De Nada\\
        <artist> Ivan Lins\\
        <lang> portuguese\\
        <schema> ABCCD\\
        <genre> MPB\\
        \\
        \text{[Generation]}\\
        Você não sabe de nada\\
        E não espera um des\\
        Vai fazer amor\\
        Depois de mais uma flor\\
        Para mim\\
        \\
        \text{[Gold]}\\
        Você não sabe de nada\\
        Se pensa que me convence\\
        As coisas que você diz\\
        São coisas que ninguém diz\\
        E isso lhe fica mal\\
        \bottomrule
    \end{tabular}}
    \caption{Multilingual generation output. The generation follows the schema even if it contains wrong words, like "des" which is just part of a word}
    \label{tab:multil;ingual_2}
\end{table}
\begin{table}[h]
\centering
\scalebox{1}{
    \begin{tabular}{p{\columnwidth}}
        \toprule
        \text{[Input]}\\
        <title> Je Suis Mes Pas\\
        <artist> Lucie Bernardoni\\
        <lang> french\\
        <schema> AAAB\\
        \\
        \text{[Generation]}\\
        Je suis mes pas\\
        Tout seul et sans voi\\
        Et moi tout bas\\
        Sans amour perdu\\
        \\
        \text{[Gold]}\\
        Le jour se lève, je brise le silence\\
        Je défie les apparences\\
        C'est le grand jour, la fin de l'innocence\\
        Il y a tant de choses à comprendre\\
        \bottomrule
    \end{tabular}}
    \caption{Multilingual output example. One of the rhymes is wrong ("voi") and moreover, it is not a real word but the start of a singular person of the verb "voir" }
    \label{tab:multilingual_3}
\end{table}

% \section{Dataset Examples}
\begin{table}[h]
    \centering
    \begin{tabular}{p{\columnwidth}}
        \toprule
        \text{[Verse 1]}\\
        I come from down in the valley\\
        Where, mister, when you're young\\
        They bring you up to do\\
        Like your daddy done\\
        Me and Mary we met in high school\\
        When she was just seventeen\\
        We'd drive out of this valley\\
        Down to where the fields were green\\
        \\
        \text{[Chorus 1]}\\
        We'd go down to the river\\
        And into the river we'd dive\\
        Oh, down to the river we'd ride\\
        % \\
        % \text{[Verse 2]}\\
        % Then I got Mary pregnant\\
        % And, man, that was all she wrote\\
        ...\\
        \bottomrule
    \end{tabular}
    \caption{Example from Genius.com data of the song "The River" by Bruce Springsteen.}
    \label{tab:the_river_full}
\end{table}

\clearpage
\subsection{Conditional generation}
\label{sec:conditional}
This subsection presents the results of our conditional generation experiment. Although the use of special tokens for conditional generation has been thoroughly studied \cite{Kprompt}, we included it to ensure our model's performance. We study how the Title and the genre affect the generated text. We used the sentence embedder $all$-$MiniLM$-$L6$-$v2$ to get the embeddings in both cases. We split the training data as $80/20$ for Train and Dev. The Test set was used to generate the synthetic data for all the models, so the values of the metrics over the Test should be considered as the reference of a human-like text. We also include the values on the Dev set to show the variance from one to another.

For the Title, we compute the dot product between the Title and the lyrics generated. Title correlation is the average of the dot product between the title embedding and the average of the verses embeddings. In the case of the genre, we trained multiple classifiers, SVM with different kernels and MPL with different structures, among others. We use Cross-Validation with a split $80/20$, obtaining the best model, SVM with linear kernel. Since the dataset has a huge imbalance, we randomly select a maximum of $700$ data points for each of the $24$ genres appearing in the test set. We use Accuracy for the genre.

The results show that conditioning generation works. The correlation between the Title and lyrics is close to the Test set, even higher in the case of the pretrained models. The genre shows a high heterogeneity among the songs from the same genre. This fact makes it complicated to obtain a good classifier. Still, the results show values very close to actual cases.

\begin{table*}[h]
    \centering
    \begin{tabular}{llcc}
        \toprule

    Model/Dataset & Decod. & Title correlation & Genre classification \\
    \toprule
    Train & Human & 0.3892 & 0.6276 \\
    Dev & Human & 0.4063 & 0.3622 \\
    Test & Human & 0.4149 & 0.2562 \\
    \midrule
    \tfbaserandom & BS & 0.2951 & 0.2460 \\
    \tfbaserandom & S+R & 0.2863 & 0.2620 \\
    \midrule
    \tflargepretrain & BS & 0.5429 & 0.2566 \\
    \tflargepretrain & S+R & 0.4092 & 0.2527 \\
    \midrule 
    \end{tabular}
    % }
    \caption{Results of the analysis in conditional generation. Genre classification is measured with the accuracy and Title correlation with the dot product. We follow the notation from \ref{sec:results} for models and decoding. T5-B indicates the base model of T5, T5-L the large version, *$_{\text{LWF}}$ indicates that the model has been trained with last-word-first, while *$^\text{Rand}$ means that it has been trained from scratch.}
    \label{tab:results_conditional}
\end{table*}

\clearpage
\subsection{Dataset statistics}
\label{sec:datasetstatistics}
More statistics on the datasets used in this work are presented below. First, in Table \ref{tab:english_dataset}, we present the statistics of the English dataset in more detail. In Table \ref{tab:multilingual_dataset_long}, we do the same for the multilingual dataset. We finish with Table \ref{tab:multilingual_datasetbylanguage} where we can find the representation of each language in the multilingual dataset.

\begin{table*}[ht!]
    \centering
    % \resizebox{\textwidth}{!}{
    \begin{tabular}{lr@{}>{\raggedleft}p{2cm}@{}>{\raggedleft}p{2cm}@{}>{\raggedleft}p{2cm}@{}>{\raggedleft}p{2cm}@{}>{\raggedleft}p{2cm}@{}>{\raggedleft}p{2cm}@{}r}
    \toprule
    Split & \# Examples & With \\ Genres & With \\ Emotions & With \\ Topics & Avg. \\Tokens & Avg. Sentences & Avg. Sentence Length &\\
    \midrule
    Train & \num{564552} & \num{146613} (25.97\%) & \num{27118} (4.80\%) & \num{170074} (30.13\%) &  46.92 & 5.84 & 8.03 &\\
    \midrule
    Dev & \num{3500} &  587 (16.77\%) &  62\\(1.77\%) & 719 (20.54\%) & 74.90 &  9.77 & 7.67&\\
    \midrule
    Test & \num{3500}  &  804 (22.97\%) & 104 \\(2.97\%) & 893 (25.51\%) &  70.79 & 9.24 & 7.67 &\\
    \bottomrule
    \end{tabular}
    % }
    \caption{Statistics for Train, Dev and Test splits of the Genius.com dataset.}
    \label{tab:english_dataset}
\end{table*}

\begin{table*}[ht!]
    \centering
    \resizebox{\textwidth}{!}{
    \begin{tabular}{lr@{}>{\raggedleft}p{2cm}@{}>{\raggedleft}p{2cm}@{}>{\raggedleft}p{2cm}@{}>{\raggedleft}p{2cm}@{}>{\raggedleft}p{2cm}@{}>{\raggedleft}p{2cm}@{}r}
    \toprule
    Split & \# Examples & With \\ Genres & With \\ Emotions & With \\ Topics & Avg. \\Tokens & Avg. Sentences & Avg. Sentence Length & Languages \\
    \midrule
    Train & \num{2588424} & \num{1238067} (47.83\%) & \num{58280} (2.25\%) & \num{170917} (6.60\%) &  40.19 & 6.99 &  5.75 & 13 \\
    \midrule
    Dev & \num{10000} &  \num{4368} &  \num{228} & \num{493} & 39.80 &  6.79 & 5.87 & 13 \\
     &  &  (43.68\%) & (2.28\%) & (4.93\%) &  &   &  &  \\
    \midrule
    Test & \num{10000} & \num{4541} & \num{211} & \num{493} &  39.80 & 6.60 & 6.03 & 13\\
     &  &  (45.41\%) & (2.11\%) & (4.93\%) &  &   &  &  \\
    \bottomrule
    \end{tabular}
    }
    \caption{Statistics for Train, Dev and Test splits of the
multilingual dataset.}
    \label{tab:multilingual_dataset_long}
\end{table*}

% \begin{table*}[ht!]
%     \centering
%     % \resizebox{\textwidth}{!}{
%     \begin{tabular}{lr@{}>{\raggedleft}p{2cm}@{}>{\raggedleft}p{2cm}@{}>{\raggedleft}p{2cm}@{}>{\raggedleft}p{2cm}@{}>{\raggedleft}p{2cm}@{}>{\raggedleft}p{2cm}@{}r}
%     \toprule
%     Split & \# Examples & With \\ Genres & With \\ Emotions & With \\ Topics & Avg. \\Tokens & Avg. Sentences & Avg. Sentence Length & Languages \\
%     \midrule
%     Train & \num{2588424} & \num{1238067} (45.06\%) & \num{58280} (2.12\%) & \num{170917} (6.22\%) &  40.19 & 6.99 &  5.75 & 15 \\
%     \midrule
%     Dev & \num{124762} &  \num{52666} &  \num{1298} & \num{742}& 39.80 &  6.79 & 5.87 & 15 \\
%      &  &  (42.21\%) & (1.04\%) & (0.59\%) &  &   &  &  \\
%     \midrule
%     Test & \num{124782} & \num{56314} & \num{1562} & \num{932} &  39.80 & 6.60 & 6.03 & 15\\
%      &  &  (45.13\%) & (1.25\%) & (0.75\%) &  &   &  &  \\
%     \bottomrule
%     \end{tabular}
%     % }
%     \caption{Statistics for Train, Dev and Test splits of the
% multilingual dataset.}
%     \label{tab:multilingual_dataset_long}
% \end{table*}
\begin{table*}[ht!]
    \centering
    % \resizebox{\textwidth}{!}{
    \begin{tabular}{lr}
    \toprule
    Language & \# Examples \\
    \midrule
    Spanish & \num{672264} \\
    English & \num{387600} \\
    German & \num{361272} \\
    French & \num{360812} \\
    Italian & \num{228386} \\
    Portuguese & \num{199367} \\
    Finnish & \num{99119} \\
    Polish & \num{86887} \\
    Dutch & \num{70605} \\
    Swedish & \num{61980} \\
    Norwegian & \num{23921} \\
    Croatian & \num{21260} \\
    Danish & \num{14951} \\
    \bottomrule
    \end{tabular}
    % }
    \caption{Statistics by language of the multilingual dataset.}
    \label{tab:multilingual_datasetbylanguage}
\end{table*}
\end{document}